*Article*

# Detecting Respiratory Pathologies Using Convolutional Neural Networks and Variational Autoencoders for Unbalancing Data

**María Teresa García-Ordás** [1,*] , **José Alberto Benítez-Andrades** [2], **Isaías García-Rodríguez** [1],
**Carmen Benavides** [2] and **Héctor Alaiz-Moretón** [1]

1   SECOMUCI Research Groups, Escuela de Ingenierías Industrial e Informática, Universidad de León, Campus de Vegazana s/n, C.P. 24071 León, Spain; isaias.garcia@unileon.es (I.G.-R.); hector.moreton@unileon.es (H.A.-M.)
2   SALBIS Research Group, Department of Electric, Systems and Automatics Engineering, Universidad de León, Campus of Vegazana s/n, 24071 León, Spain; jbena@unileon.es (J.A.B.-A.); mcbenc@unileon.es (C.B.)
*   Correspondence: mgaro@unileon.es

**Abstract:** The aim of this paper was the detection of pathologies through respiratory sounds. The ICBHI (International Conference on Biomedical and Health Informatics) Benchmark was used. This dataset is composed of 920 sounds of which 810 are of chronic diseases, 75 of non-chronic diseases and only 35 of healthy individuals. As more than 88% of the samples of the dataset are from the same class (Chronic), the use of a Variational Convolutional Autoencoder was proposed to generate new labeled data and other well known oversampling techniques after determining that the dataset classes are unbalanced. Once the preprocessing step was carried out, a Convolutional Neural Network (CNN) was used to classify the respiratory sounds into healthy, chronic, and non-chronic disease. In addition, we carried out a more challenging classification trying to distinguish between the different types of pathologies or healthy: URTI, COPD, Bronchiectasis, Pneumonia, and Bronchiolitis. We achieved results up to 0.993 F-Score in the three-label classification and 0.990 F-Score in the more challenging six-class classification.

## 1. Introduction

Today, respiratory pathologies are a common problem all over the world. Although smoking is the most common cause of respiratory pathologies, sometimes they are caused by genetics, as well as environmental exposure [1]. The ICBHI (International Conference on Biomedical and Health Informatics) respiratory dataset [2] includes seven pathologies, such as chronic obstructive pulmonary disease (COPD), asthma, upper respiratory tract infection (URTI), lower respiratory tract infection (LRTI), bronchiectasis, pneumonia, and bronchiolitis.

Chronic obstructive pulmonary disease (COPD) is a chronic pathology which is difficult to detect. The main cause of COPD is smoking [3]. It causes symptoms, including shortness of breath and cough, which are also common in Asthma disease. Furthermore, these symptoms can be interpreted as a simple aging process.

Bronchiectasis is a chronic condition in which the airways of the lungs become abnormally widened. These damaged air passages allow bacteria and mucus to build up and pool in your lungs. This results in frequent infections and blockages of the airways. All these symptoms can be interpreted also as a bronchiolitis or just a cold [4]. The main difference between both diseases is that bronchiolitis most often affects young children and it can be cure, whereas bronchiectasis is a chronic disease.

Upper respiratory tract infection is a non-chronic disease that can happen at any time, but it is more common in the fall and winter. The vast majority of upper respiratory infections are caused by viruses [5]. The symptoms of this disease can be confused with those of pneumonia [1]. Most people with pneumonia can recover in a short time, but for certain people, it can be extremely serious and even life-threatening so the diagnosis is crucial.

Symptoms of lower respiratory tract infections (LRTI) vary and depend on the severity of the infection. Less severe infections can have symptoms similar to those of bronchiectasis or bronchiolitis.

As we can see, the symptoms of all these diseases are very common and can cause a bad diagnosis by the doctor. For all this, it is very interesting to be able to determine the disease using the sound of the breaths without taking into account the rest of the symptoms.

## 2. Related Work

### 2.1. Respiratory Sounds Detection

Lung auscultation provides valuable information regarding the respiratory function of the patient, and it is important to analyze respiratory sounds using an algorithm to give support to medical doctors. There are a few methods in the literature to deal with this challenge. Typically, wheezing is found in asthma and chronic obstructive lung diseases. Wheezes can be so loud you can hear it just by standing next to the patient. Crackles, on the other hand, are only heard using a stethoscope, and they are a sign of too much fluid in the lung. Crackles and wheezes are indications of the pathology.

Islam et al. [6] detected asthma pathology by basing their research on the fact that asthma detection from lung sound signals rely on the presence of wheeze. They collected lung sounds from 60 subjects in which the 50% had asthma and using a data acquisition system from four different positions on the back of the chest. For the classification step, ANN (Artificial Neural Networks) and SVM (Support Vector Machine) were used with the best results (93.3%) obtained in the SVM scenario.

Other studies based on the detection of wheezes and crackles [7] used different configurations of a neural network, obtaining results of up to 93% for detecting crackles and 91.7% for wheezes. The same goal is pursued in Reference [8], but the dataset used in this case consists of seven classes: normal, coarse crackle, fine crackle, monophonic wheeze, polyphonic wheeze, squawk, and stridor. The best results were achieved using a Convolutional Neural Network (CNN). Chen et al. [9] proposed ResNet with an OST-based (Optimized S-Transform based) feature map to classify wheeze, crackle, and normal sounds. In detail, three RGB -maps (Red-Green-Blue) of the rescaled feature map is fed into ResNet due to the balance between the depth and performance. The input feature map is passed through three steps of the ResNet structure and finally, the output corresponds to the class (wheeze, crackle, or normal). The results are compared with ResNet-STFT (Short Term Fourier Transform) and ResNet-ST (S-Transform), with the best accuracy achieved using their proposal ResNet-OST.

In Reference [10], the authors propose a methodology to classify the respiratory sounds into wheezes, crackles, both wheezes and crackles, and normal using the same dataset as that used in our work: ICBHI [2]. The procedure consists of a noise suppression step using spectral subtraction followed by a feature extraction process. Hidden Markov Models were used in the classification step obtaining 39.56% using the score metric, defined as the average of sensitivity and specificity. These results are not promising but in Reference [11], Perna et al. proposed a reliable method to classify in healthy, chronic disease, or non-chronic disease based-on wheezes, crackles, or normal sounds using deep learning and, more concretely, recurrent neural networks and again using the ICBHI benchmark [2].

In Reference [12], Jacome et al. also proposed a CNN to deal with respiratory sounds for detecting breathing phase with a 97% of success in inspiration detection and a 87% in expiration.

Early models of RNNs suffered from both exploding and vanishing gradient problems. Long Short Term Memory (LSTM) and Gated Recurrent Unit (GRU) were designed to address the gradient problems successfully. The authors exploited the LSTM and GRU advantages and obtained promising results of up to 91 % of the ICBHI Score calculated as the average value of sensitivity and specificity [11].

Deep learning techniques have also been used to detect some kinds of pathologies such as bronchiolitis, URTI, pneumonia, etc., which supposed a more challenging problem than classifying wheezes and crackles. In Reference [13], the authors try to distinguish between pathological and non-pathological voice over the Saarbrücken Voice Database (SVD) using the MultiFocal toolkit for a discriminative calibration and fusion. The authors carry out a feature extraction step, and these features (Mel-frequency cepstral coefficients, harmonics-to-noise ratio, normalized noise energy and glottal-to-noise excitation ratio) are used to train a generative Gaussian Mixture Models (GMM) model [14].

## 2.2. Deep Learning Techniques

In the literature, many deep learning techniques have been used to resolve all kinds of problem. This, gives us an idea of how useful artificial intelligence is.

Today, one of the most used techniques for all kinds of purposes are autoencoders and CNNs. Sugimoto et al. [15] try to detect myocardial infarction using the ECG (electrocardiogram) information using a CNN. In their experiments, the classification performance was evaluated using 353,640 beats obtained from the ECG data of MI (myocardial infarction) patients and healthy subjects. ECG data was extremely imbalanced, and the minority class, including abnormal ECG data, may not be learned adequately. To solve this problem, the authors proposed to use the convolutional autoencoder in the following way: The CAE model is constructed for each lead and outputs reconstructed input ECG data if normal ECG data is inputted. Otherwise, the waveform is distorted and outputted. After this process, k-Nearest Neighbors (kNN) is used as a classifier.

A CAE (Convolutional autoencoder) is also used in Reference [16] to restore the corrupted laser stripe images of the depth sensor by denoising the data.

In Reference [17], Kao et al. propose a method of classifying Lycopersicons based on three levels of maturity (immature, semi-mature, and mature). Their method includes two artificial neural networks, a convolutional autoencoder (CAE), and a backpropagation neural network. With the first one, the ROI in the Lycopersicon is detected (instead of doing it manually). Then, using the extracted features, the neural network employs self-learning mechanisms to determine Lycopersicon maturity obtaining an accuracy rate of 100%.

A variational autoencoder is used in Reference [18] for video anomaly detection and localization using only normal samples. The method is based on Gaussian Mixture Variational Autoencoder, which can learn the feature representations of the normal samples as a Gaussian Mixture Model trained using deep learning. A Fully Convolutional Network (FCN) is employed for the encoder-decoder structure to preserve relative spatial coordinates between the input image and the output feature map.

In Reference [19], a non-linear surrogate model based on deep learning is proposed using a variational autoencoder with deep convolutional layers and a deep neural network with batch normalization (VAEDC-DNN) for a real-time analysis of the probability of death in toxic gas scenarios.

Advances in indoor positioning technologies can generate large volumes of spatial trajectory data on the occupants. These data can reveal the distribution of the occupants. In Reference [20], the authors propose a method of evaluating similarities in occupant trajectory data using a convolutional autoencoder (CAE).

In Reference [21], deep autoencoders are proposed to produce efficient bimodal features from the audio and visual stream inputs. The authors obtained an average relative reduction of 36.9% for a range of different noisy conditions, and also, a relative reduction of 19.2% for the clean condition in terms of the Phoneme Error Rates (PER) in comparison with the baseline method.

In 2019, variational autoencoders have been widely used to analyze different kind of signals and monitoring them [22,23]. In addition, in Zemouri et al. [24], variational autoencoders have been used for train a model as a 2D visualization tool for partial discharge source classification.

So, today autoencoders and CNNs are widely used in the literature to solve all kinds of problems. We take advantage of both methods and proposed the use of a variational convolutional autoencoder to balance the data, as well as a CNN to carry out the classification step.

For this paper, we proposed a technique to classify healthy, chronic disease, and non-chronic disease and six different pathology classes: Chronic obstructive pulmonary disease (COPD), upper respiratory tract infection (URTI), bronchiectasis, pneumonia, bronchiolitis, and healthy. Our procedure outperforms the state-of-the-art proposals.

The rest of the paper is organized as follows: In Section 3, we describe the methodology, including data preprocessing, data normalization, and data augmentation, using our Variational Convolutional autoencoder and finally data classification using a CNN. Experiments and results are detailed in Section 4, taking into account two types of classification, and finally, we conclude in Section 5.

## 3. Methods and Materials

### 3.1. Data Normalization

Data normalization is an important step before carrying out any machine learning strategy. There are multiple alternatives to normalize data. In this paper, we evaluated our data with MinMax normalization, which got the data in the [0,1] range.

### 3.2. Data Augmentation: Variational AutoEncoder (VAE)

In all fields of research, but more frequently in ehealth, it is very common to have unbalanced data in the datasets. That means that the number of elements (cardinal) of one class is much bigger than all the cardinal of the rest of the classes. To solve this, there are multiples techniques which try to replicate samples of the minority classes. Synthetic Minority Oversampling Technique (SMOTE) [25], Adaptive Synthetic Sampling Method (ADASYN) [26], and Variational Autoencoders (VAE) [27] are some examples of generative methods. We evaluated our dataset with all of these oversampling methods obtaining the best results with VAE, as shown in the results section.

VAE are part of a kind of neural network known as autoencoders. Vanilla autoencoder architecture consists of a number of dense hidden layers with two main peculiarities:

- One of these hidden layers has a very few neurons (latent space).
- The output of the vanilla autoencoder tries to replicate the input.

Taking into account these two considerations, when an autoencoder is trained, the net learns to encode data information in its latent space and decode them after that to reconstruct the original data.

However, VAE is a probabilistic model focused on learning the distribution of data to be able to create new samples which would belong to this distribution. Whereas Vanilla autoencoders try to reconstruct the original data, VAE is also trained to learn the distribution of the data. For that reason, the loss function used to train a VAE is made up of two terms: "reconstruction term", like in the vanilla autoencoder, that tends to make the encoder-decoder work accurately; and a "regularization term" applied over the latent layer that tends to make the distributions created by the encoder close to a standard normal distribution using the Kulback-Leibler divergence (see Equation (1)).

$$VAE\_LOSS = ||x - \bar{x}||^2 + KL[N(\mu_x, \sigma_x), N(0,1)], \qquad (1)$$

where $\bar{x}$ is the reconstruction of $x$, and $N(\mu_x, \sigma_x)$ a normal distribution with mean $\mu_x$ and standard deviation $\sigma_x$ and $KL[p, q]$ is the Kulback-Leilber divergence defined in Equation (2):

$$KL[p, q] = -\int p(x) \log q(x) dx + \int p(x) \log p(x) dx. \qquad (2)$$

In Figure 1a,b, respectively, we show the differences between a variational autoencoder and a vanilla one.

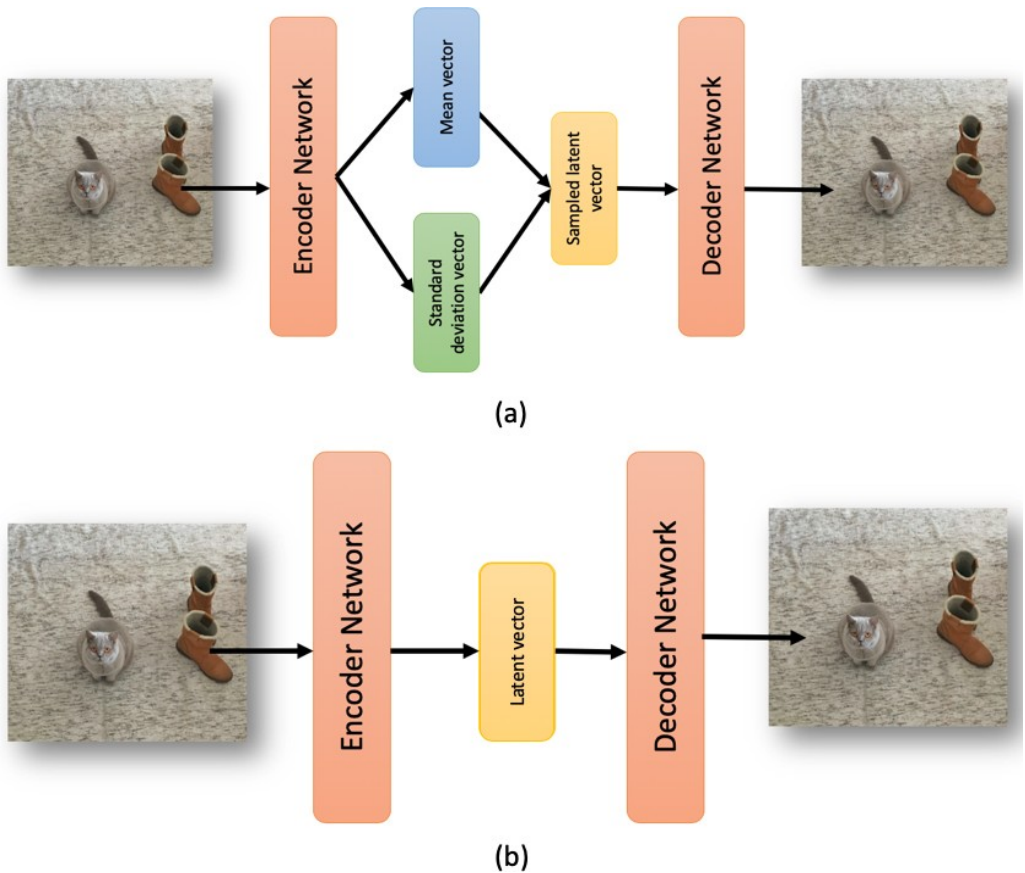

**(a)**

**(b)**

**Figure 1.** In (**a**), a Variational AutoEncoder (VAE) scheme with the mean and standard deviation layers used to sample the latent vector. In (**b**), the vanilla autoencoder with a simple latent vector.

### 3.3. Data Classification: Convolutional Neural Networks

A CNN is a Deep Learning algorithm which can take in a bi-dimensional input and be able to differentiate it from another by learning filters which extracts complex features from the inputs automatically. A basic modeling of a CNN is represented in Figure 2.

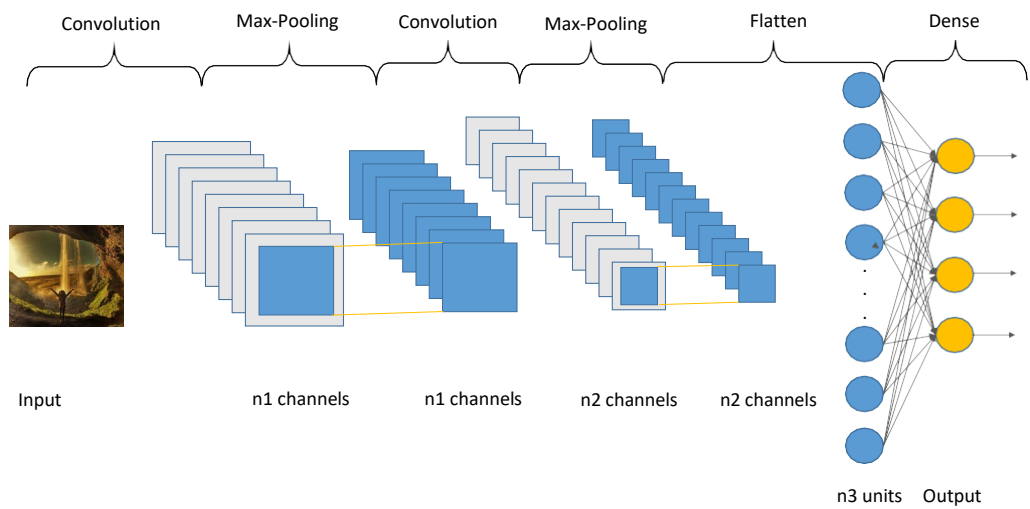

**Figure 2.** A vanilla Convolutional Neural Network (CNN) representation.

During the training step, each convolution layer learns the filter weights to then produce a feature map. The filter or kernel is sliding over the input and the sum of the convolution generates the feature map.

After a convolution layer, it is common to add a pooling layer. These kinds of layer are use to decrease the number of parameters in the network. This reduces the computational cost and controls overfitting. The most frequent type of pooling is Max-pooling, which takes the maximum value in each window. In order to carry out a classification or a regression problem with the features generated by the convolutional layers, it is necessary to add dense layers at the end of the network.

## 4. Experiments and Results

### 4.1. Dataset

The ICBHI (International Conference on Biomedical and Health Informatics) dataset [2] was created by two research teams (Greece and Portugal), and it includes 920 recordings acquired from 126 subjects. A total of 6898 respiration cycles and 5.5 h of sound was recorded. One thousand, eight hundred and sixty-four of these 6898 respiration cycles were labeled as crackles; 886 contain wheezes and 506 contain both crackles and wheezes. Crackles and wheezes were labeled by experts in the field. Respiratory sounds were recorder from seven different chest locations: trachea, left and right anterior, left and right posterior, and left and right lateral (see Figure 3). High noise levels were included to simulate real situations obtaining a challenging dataset.

Respiratory sounds were recorded from patients with chronic obstructive pulmonary disease (COPD), asthma, upper respiratory tract infection (URTI), lower respiratory tract infection (LRTI), bronchiectasis, pneumonia, and bronchiolitis.

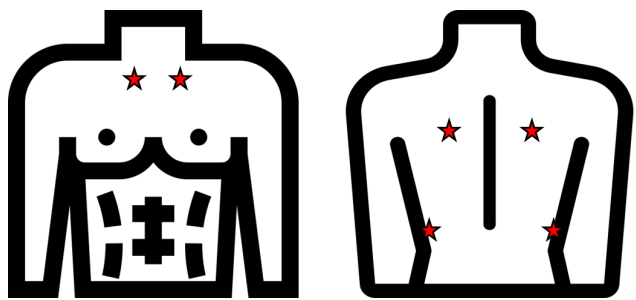

**Figure 3.** The sounds were recorded from seven different locations remarked in red.

### 4.1.1. Image Generation

In this paper, we are going to deal with audios using their Mel Spectrogram. A Mel Spectogram is a visual representation of the spectrum of a sound on the Mel scale. Mel Scale [28], proposed by Stevens et al. is a perceptual scale of equally-spaced pitches. The conversion of hertzs into Mels is done using Equation (3).

$$m = 2595 \times \log_{10}\left(1 + \frac{f}{700}\right),\tag{3}$$

where *f* is the frecuency in hertzs.

There are four steps to be carried out to obtain the Mel spectrogram given an audio input:

1. Sampling the input wave with windows of a fixed size and step.
2. Compute the Fast Fourier Transform to get the data to the frequency domain.
3. Generate bins using the Mel scale.

4.  Generate spectogram breaking down the magnitude of the signal into the frequencies of the Mel scale.

After all the spectrograms are built, all the images were resized to have the same number of columns. Each column represents a unit of time, so it is very common to have different sizes throughout all of our datasets. In our case, all the images were resized to the mean number of columns in all the spectrograms (see Equation (4)).

$$mean\_columns = \sum_{i=1}^{N} \frac{cols(dataset(i))}{N},　　　　　　(4)$$

where $N$ is the number of spectrograms in the experiment and $cols(x)$ the number of columns of image $x$. Some of these images can be seen in Figure 4.

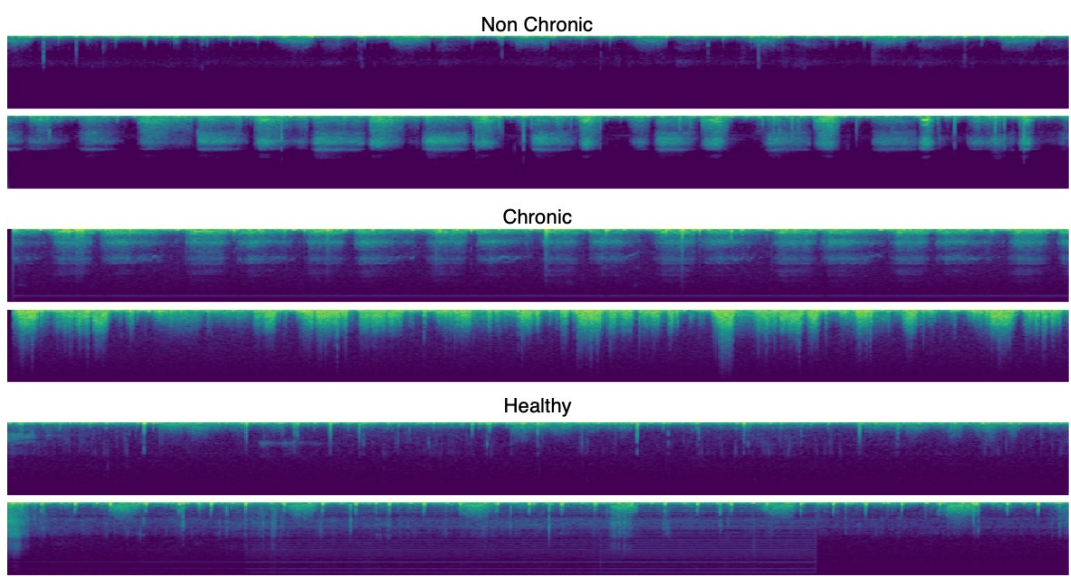

**Figure 4.** Examples of the Mel Spectrograms for the chronic, non-chronic and healthy classes after preprocessing.

### 4.2. Chronic Classification

#### 4.2.1. Experimental Setup

In this work, a Min-Max feature normalization was carried out to set our data in the range of [0,1] which highly improves the performance of the neural network training. After that, we evaluated the class distribution of the dataset taking into account three different values: Chronic, Non-Chronic, and Healthy. In Table 1, we can see the unbalanced distribution.

**Table 1.** Number of samples for each class.

|  | # |
| --- | --- |
| Chronic | 810 |
| Non-Chronic | 75 |
| Healthy | 35 |

As we can see, the number of samples of chronic pathologies represents 88.04% of all the dataset. In our experiments, we carried out a classification on the unbalanced dataset with and without using class weights. Furthermore, an augmentation of the less representative classes was done to balance the dataset. This augmentation step was carried out using our proposed VAE.

A convolutional VAE scheme has been implemented in order to generate more samples for the Non-Chronic and healthy classes. In Figure 5, we can see the network configuration.

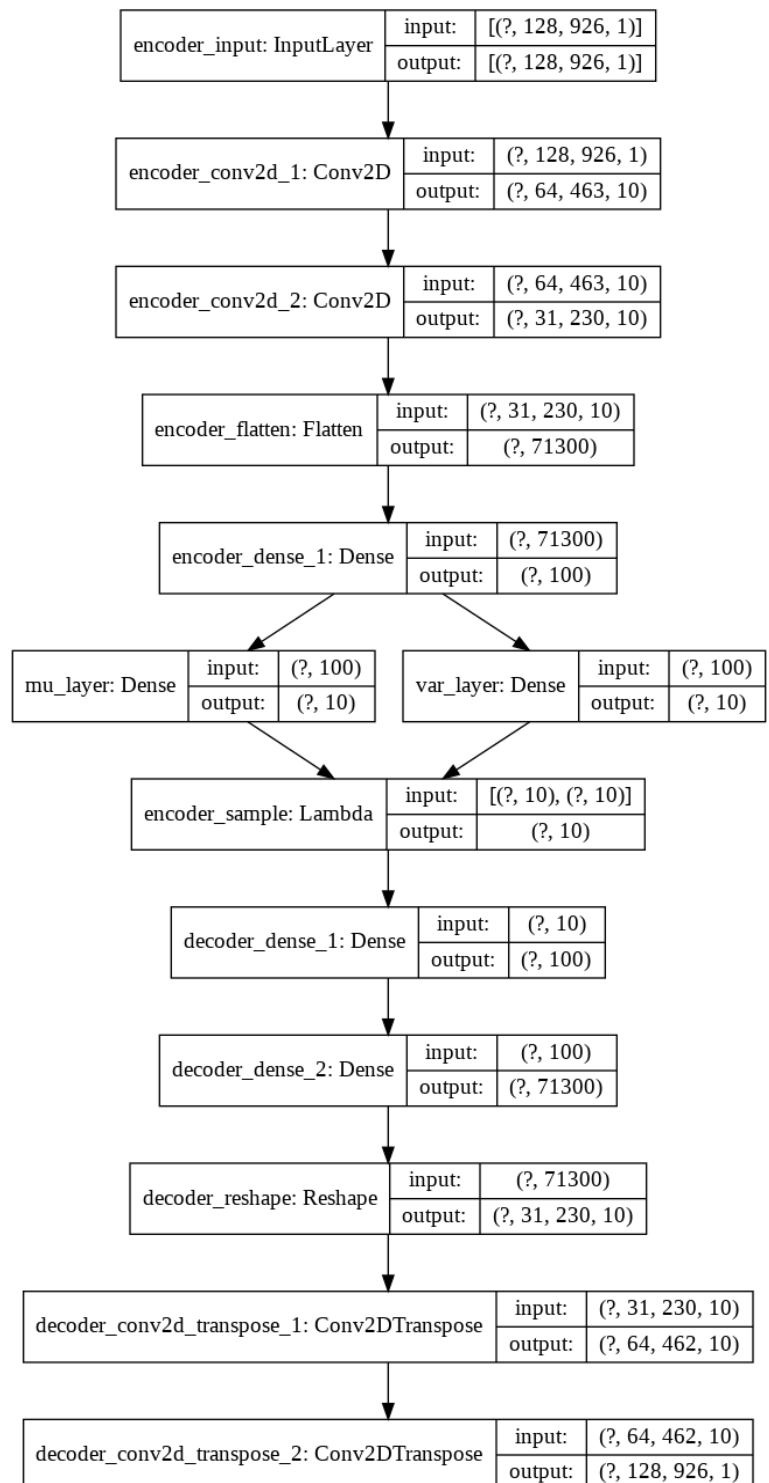

**Figure 5.** VAE scheme configuration for data augmentation.

In Table 2, the new size of each class is shown.

**Table 2.** Number of samples for each class after data augmentation.

|  | # |
|---|---|
| Chronic | 810 |
| Non-Chronic | 900 |
| Healthy | 840 |

Some examples of the new images generated can be seen in Figure 6.

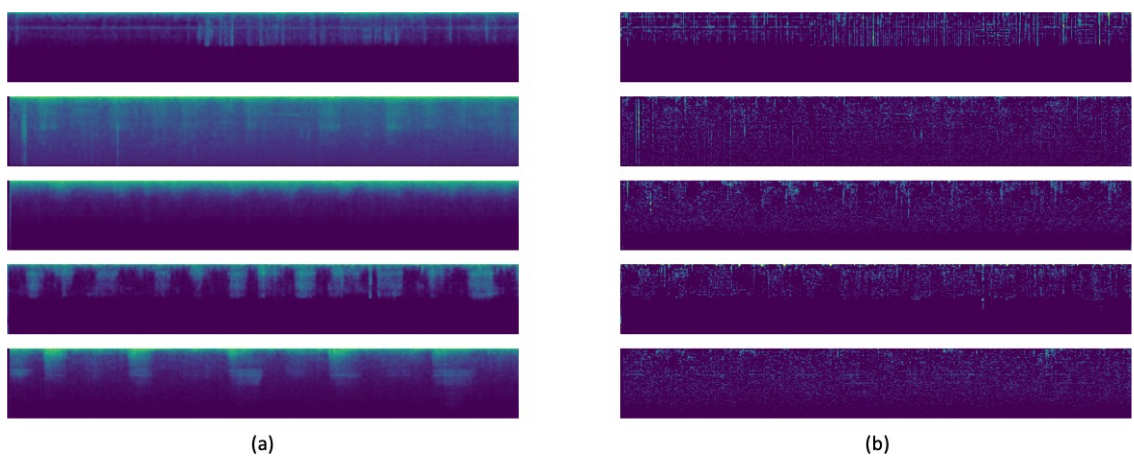

(a)　　　　　　　　　　　　　　　　　　　　　　　　(b)

**Figure 6.** In (**a**), the new images generated using the VAE. In (**b**), the variation between the original and the generated images.

Once we have our dataset well balanced, we designed a CNN for three class specifications. The scheme of this network can be seen in Figure 4. As we can see, we added some layers such as BatchNormalization and Dropout to avoid the overfitting problem. The output layer consists of three neurons to fit the three class classification.

We used Adam as the optimization algorithm and categorical crossentropy as the loss function. Before training, a train-test split (80-20) was carried out in order to clearly distinguish between the data used for training and that used to evaluate the classifier.

In order to avoid random factor, a 10 cross validation has been carried out in all the experiments showing in all the tables the mean value of the metrics. The intermediate results for the proposal can be shown in Table 3.

**Table 3.** All the iteration results for the 10 cross validation step using the proposal combination of Variational AutoEncoder (VAE) for data augmentation and a CNN for the classification step on chronic diseases detection.

|  | 1 | 2 | 3 | 4 | 5 | 6 | 7 | 8 | 9 | 10 | Mean | Std |
|---|---|---|---|---|---|---|---|---|---|---|---|---|
| **Sensitivity** | 0.982709 | 0.985591 | 0.988473 | 0.985591 | 0.988473 | 0.979827 | 0.991354 | 0.976945 | 0.988473 | 0.982709 | **0.985014** | 0.004465 |
| **Specificity** | 0.987730 | 0.993865 | 0.987730 | 0.987730 | 0.987730 | 0.993865 | 0.993865 | 0.987730 | 0.987730 | 0.993865 | **0.990184** | 0.003168 |
| **Score** | 0.985219 | 0.989728 | 0.988101 | 0.986660 | 0.988101 | 0.986846 | 0.992610 | 0.982338 | 0.988101 | 0.988287 | **0.987599** | 0.002701 |
| **Precision** | 0.993827 | 1.000000 | 1.000000 | 0.993827 | 0.993827 | 0.993865 | 0.993865 | 0.993827 | 1.000000 | 1.000000 | **0.996304** | 0.003181 |
| **Recall** | 0.987730 | 0.993865 | 0.987730 | 0.987730 | 0.987730 | 0.993865 | 0.993865 | 0.987730 | 0.987730 | 0.993865 | **0.990184** | 0.003168 |
| **FScore** | 0.990769 | 0.996923 | 0.993827 | 0.990769 | 0.990769 | 0.993865 | 0.993865 | 0.990769 | 0.993827 | 0.996923 | **0.993231** | 0.002427 |

As we can see, the standard deviation is very small, which indicates the good generalization of the method using different dataset splits.

### 4.2.2. Chronic Classification Results

We carried out five different classifications using the CNN scheme shown in Table 4.

**Table 4.** Scheme of the classification neural network based on convolutional layers.

| Number of Layer | Name Layer | Type | Input Shape | Output Shape |
|---|---|---|---|---|
| 1 | cnn_input | InputLayer | [(?,128,926,1)] | [(?,128,926,1)] |
| 2 | cnn_conv2D | Conv2D | [(?,128,926,1)] | [(?,126,924,10)] |
| 3 | cnn_batch_norm | BatchNormalization | [(?,126,924,10)] | [(?,126,924,10)] |
| 4 | cnn_activation_relu | Activation | [(?,126,924,10)] | [(?,126,924,10)] |
| 5 | cnn_dropout | Dropout | [(?,126,924,10)] | [(?,126,924,10)] |
| 6 | cnn_maxpooling2d | MaxPooling2D | [(?,126,924,10)] | [(?,25,184,10)] |
| 7 | cnn_flatten | Flatten | [(?,25,184,10)] | [(?,46000)] |
| 8 | cnn_dense | Dense | [(?,46000)] | [(?,100)] |
| 9 | cnn_output | Dense | [(?,100)] | [(?,3)] |

The first two classifications were made without a data augmentation process, which led to a very unbalanced training. In the first experiment, we trained our data without modifications, while for the second, we calculated the training weights for each class based on their number of elements. The rest of the classification were made by adding the new elements generated with SMOTE, ADASYN, and our convolutional VAE network to the data set.

We used Sensitivity (Recall), Specificity, and Score metrics defined in the same way as the authors did in Reference [11]:

$$Sensitivity = \frac{C_{chronic} + C_{non-chronic}}{N_{chronic} + N_{non-chronic}}, \tag{5}$$

$$Specificity = \frac{C_{healthy}}{N_{healthy}}, \tag{6}$$

$$Score = \frac{Sensitivity + Specificity}{2}, \tag{7}$$

where $C$ represents the correctly recognize samples, and $N$ represents the total of the specified class.

In Table 5, we can see the metrics obtained with the five experiments over the chronic classification. Furthermore, well-known metrics, such as precision, recall, and F-Score, have been calculated.

**Table 5.** Metric results obtained with the five classifications over chronic datasets.

|  | Sensitivity | Specificity | Score | Precision | Recall | F-Score |
|---|---|---|---|---|---|---|
| Dataset unbalanced | 0.941 | 0 | 0.471 | 0 | 0 | 0 |
| Dataset weighted | 0.953 | 0 | 0.476 | 0 | 0 | 0 |
| **Dataset VAE** | **0.985** | **0.990** | **0.988** | **0.996** | **0.990** | **0.993** |
| Dataset SMOTE | 0.950 | 0.167 | 0.558 | 0.500 | 0.167 | 0.250 |
| Dataset ADASYN | 0.965 | 0.857 | 0.911 | 0.857 | 0.857 | 0.857 |

The dataset oversampled using VAE achieved the best results with all the metrics. Whereas sensitivity has very high values in all the cases, specificity, precision, and recall show a very poor performance in all the other cases due to the miss-classification of the healthy class. It is also important to notice the high classification of healthy individuals according to the precision score. This is very important due to the high risk of classify a non-chronic or, even worse, a chronic disease as healthy.

In Figure 7, the confusion matrix obtained for the unbalanced classifications and the best oversampling technique are shown.

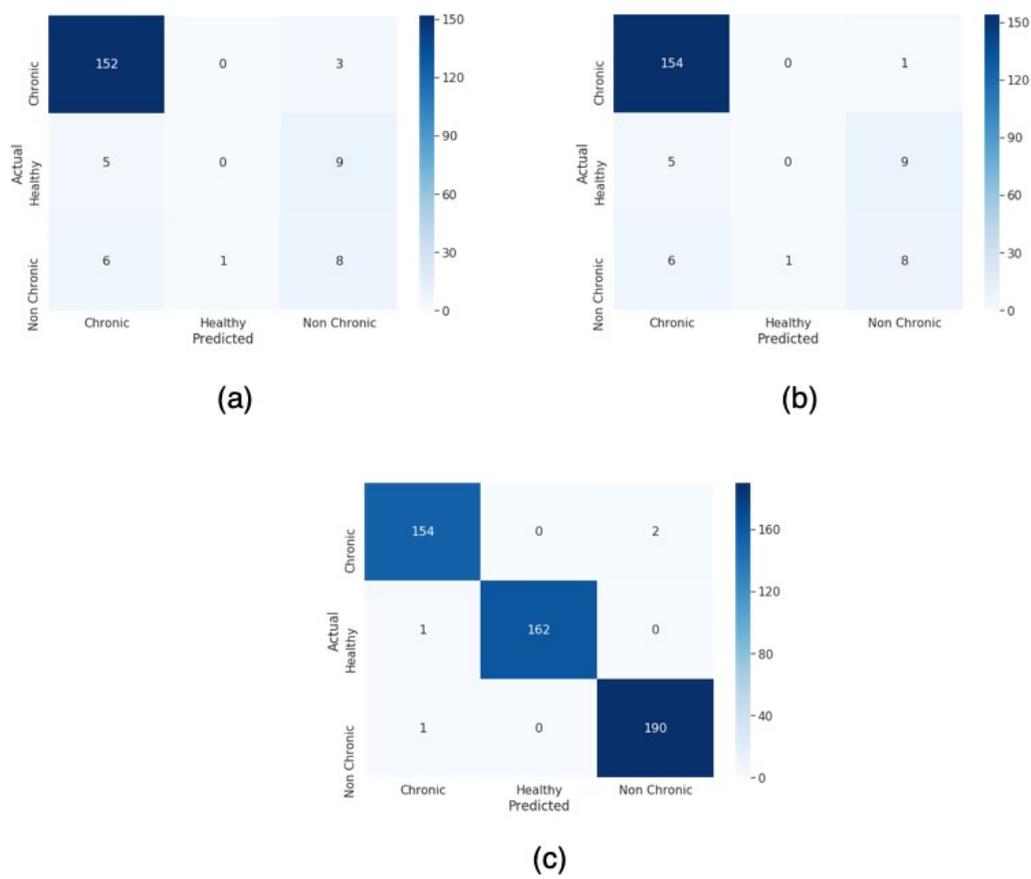

**Figure 7.** (**a**) Confusion matrix of the unbalanced dataset. (**b**) Confusion matrix of the unbalanced dataset with weights in the training. (**c**) Confusion matrix of balanced dataset using our proposal scheme.

As we can see, in all the experiments, the Chronic class obtains the better results. However, the unbalanced experiments show a very bad performance in the healthy classification, obtaining 0% of Specifity. Our proposal with the balanced dataset demonstrate that by adding new synthetic Mel Spectrogram created with a convolutional VAE yields very good classification for all the classes.

Furthermore, in Figure 8, we can see a comparison between our proposal and the methods found in the state-of-the-art of the ICBHI dataset. The results show that our method improves all of the papers in the state-of-the-art.

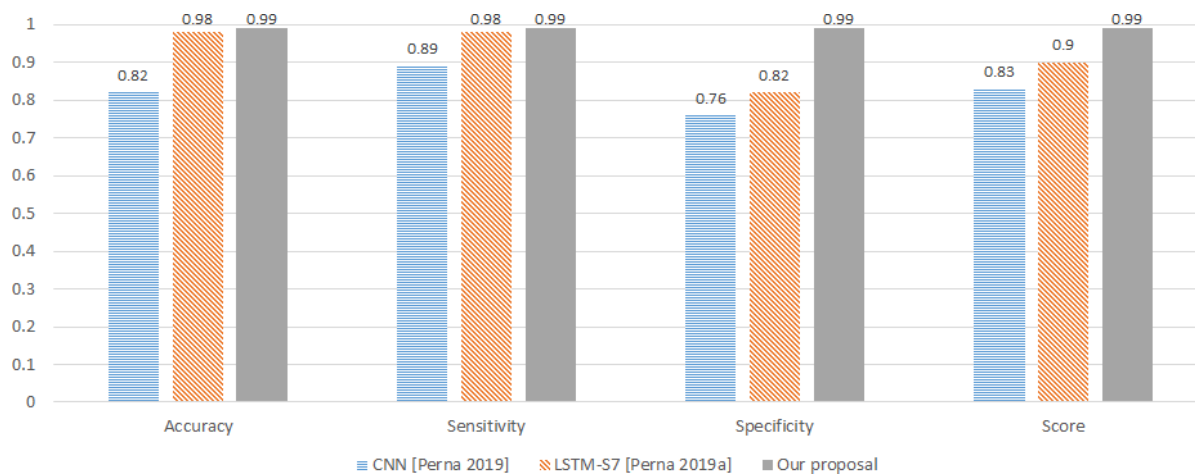

**Figure 8.** Comparative between our proposal method and the best results on the state-of-the-art using (International Conference on Biomedical and Health Informatics (ICBHI) dataset.

### 4.3. Pathology Classification

#### 4.3.1. Experimental Setup

As we did in the chronic classification, the first step we carried out was a normalization using the min-max scaler techinque. A study of the distribution of the pathology classes was done, and it demonstrates that 86.20% of the samples belong to COPD disease (see Table 6).

**Table 6.** Number of samples for each class.

|  | # |
|---|---|
| COPD | 793 |
| Pneumonia | 37 |
| Healthy | 35 |
| URTI | 23 |
| Bronchiectasis | 16 |
| Bronchiolitis | 13 |
| LRTI | 2 |
| Asthma | 1 |

LRTI and asthma have just two and one samples, respectively, so we decided to ignore them for our classification. The same augmentation data algorithm has been carried out to increase the number of samples of all the diseases except that of COPD. In the end, our dataset was made up of a total amount of 4874 spectrograms. We used the same CNN as in Section 4.2.1, except for the output layer which, in this case, has six neurons, one for each class. All of the training parameters were also the same.

In order to avoid random factor, a 10 cross validation has been carried out in all the experiments showing in all the tables the mean value of the metrics. The intermediate results for the proposal can be shown in Table 7.

**Table 7.** All the iteration results for the 10 cross validation steps using the proposal combination of VAE for data augmentation and a CNN for the classification step on pathologies dataset.

| | 1 | 2 | 3 | 4 | 5 | 6 | 7 | 8 | 9 | 10 | Mean | Std |
|---|---|---|---|---|---|---|---|---|---|---|---|---|
| **Sensitivity** | 0.983234 | 0.992528 | 0.992701 | 0.982885 | 0.983811 | 0.983071 | 0.990025 | 0.991515 | 0.995110 | 0.986420 | **0.988130** | 0.004745 |
| **Specificity** | 0.992857 | 0.994186 | 0.986928 | 0.980892 | 0.982558 | 1.000000 | 0.994220 | 0.973333 | 0.980892 | 0.975758 | **0.986162** | 0.008867 |
| **Score** | 0.988045 | 0.993357 | 0.989814 | 0.981888 | 0.983184 | 0.991536 | 0.992122 | 0.982424 | 0.988001 | 0.981089 | **0.987146** | 0.004638 |
| **Precision** | 0.992857 | 1.000000 | 0.980519 | 0.987179 | 1.000000 | 0.986667 | 0.994220 | 1.000000 | 1.000000 | 0.993827 | **0.993527** | 0.006873 |
| **Recall** | 0.992857 | 0.994186 | 0.986928 | 0.980892 | 0.982558 | 1.000000 | 0.994220 | 0.973333 | 0.980892 | 0.975758 | **0.986162** | 0.008867 |
| **FScore** | 0.992857 | 0.997085 | 0.983713 | 0.984026 | 0.991202 | 0.993289 | 0.994220 | 0.986486 | 0.990354 | 0.984709 | **0.989794** | 0.004757 |

As we can see, the standard deviation is also very small on pathologies dataset, which indicates the good generalization of the method using different dataset splits.

### 4.3.2. Pathology Classification Results

As we did in the chronic classification, we carried out five different classifications using the CNN scheme shown in Table 4 for unbalanced, weighted and balanced datasets. For the classification of most pathologies, the more challenging one, the same metrics were defined in the following way:

$$Sensitivity = \frac{C_{URTI} + C_{COPD} + C_{Bronchiectasis} + C_{Pneumonia} + C_{Bronchiolitis}}{N_{healthy} + N_{non-healthy}}, \tag{8}$$

$$Specificity = \frac{C_{healthy}}{N_{healthy}}, \tag{9}$$

$$Score = \frac{Sensitivity + Specificity}{2}. \tag{10}$$

In Table 8, the aforementioned metrics were calculated for each experiment.

**Table 8.** Metric results obtained with the three classifications over pathologies datasets.

| | Sensitivity | Specificity | Score | Precision | Recall | F-Score |
|---|---|---|---|---|---|---|
| Dataset unbalanced | 0.888 | 0.286 | 0.587 | 0.444 | 0.288 | 0.349 |
| Dataset weighted | 0.912 | 0.071 | 0.492 | 0.250 | 0.071 | 0.111 |
| **Dataset VAE** | **0.988** | **0.986** | **0.987** | **0.994** | **0.986** | **0.900** |
| Dataset SMOTE | 0.876 | 0.429 | 0.653 | 0.545 | 0.429 | 0.480 |
| Dataset ADASYN | 0.917 | 0.667 | 0.792 | 0.500 | 0.667 | 0.571 |

As we can see in the table, the behavior is exactly the same as in the chronic-non-chronic detection. Dataset augmented using our VAE architecture proposal, outperforms with all the metrics the performance shown by all the other augmentation techniques and even with the raw dataset. In comparison with the ternary classification, on this experiment just VAE augmented dataset achieved optimal results according to Sensitivity. Taking into account the F-Score value, which is one of the most reliable metrics, of our proposal outperforms the second better method (ADASYN augmentation) by more than 72%.

In Figure 9, the three confusion matrix obtained for the two unbalanced and the best balanced dataset are shown.

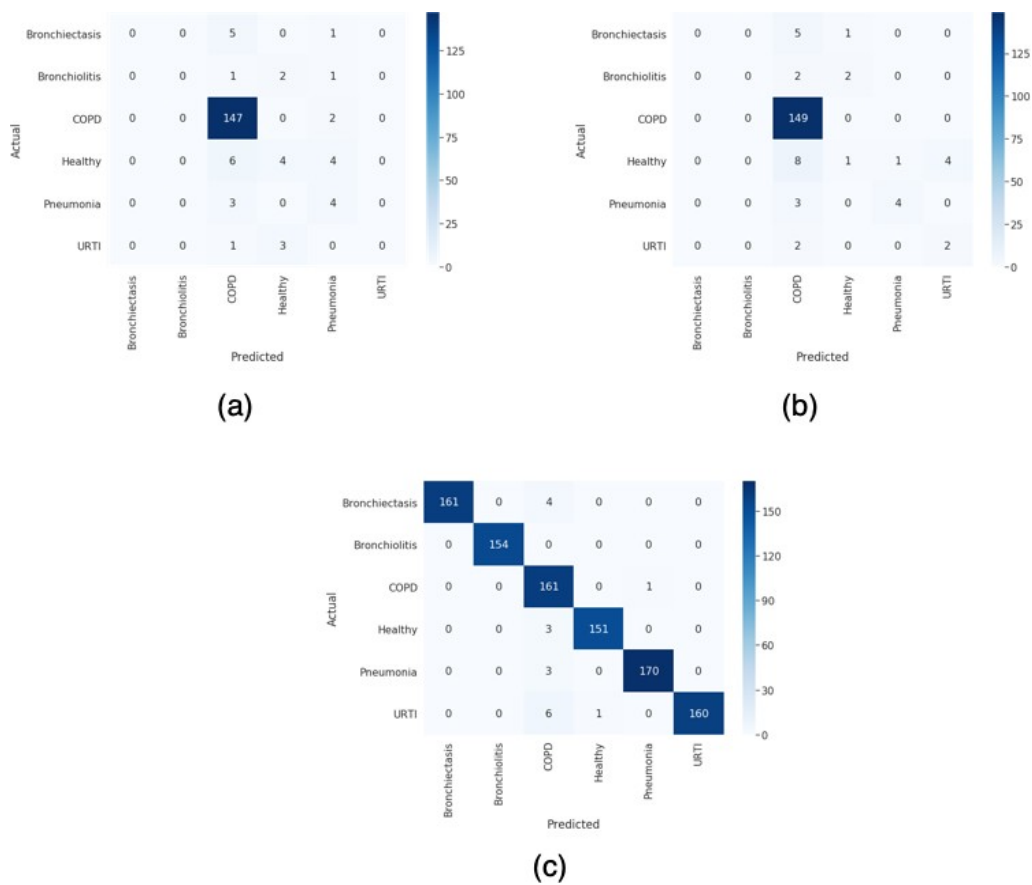

**Figure 9.** (**a**) Confusion matrix of the unbalanced dataset. (**b**) Confusion matrix of the unbalanced dataset with weights in the training. (**c**) Confusion matrix of the balanced dataset using our proposed scheme with VAE.

## 5. Conclusions and Future Work

In this article, a new procedure has been proposed to detect respiratory pathologies. In the analysis of medical data, it is very common to have very unbalanced data sets. In our work, we proposed a convolutional variational autoencoder to increase the rare classes. We transformed all respiratory audios into Mel Spectrograms to work with convolutional networks. These types of networks have very fast learning using GPUs and can learn the most relevant characteristics of the images analyzed for themselves without the need for a description step. We carried out two different locations; one for detecting chronic, non-chronic, and healthy pathologies in breathing and the other for identifying these pathologies from each other. In the first experiment, results showed 0.991 of sensitivity and a 0.994 of specifity outperforming all the studies of the state-of-the-art. Furthermore, a new and more challenging experiment with five different (and the healthy one) classes was carried out, with promising results, with the same CNN with a 0.988 score in sensitivity metric and a 0.986 in specificity.

With these results, we can conclude that using Mel Spectrograms and CNNs, pathologies in sounds of breaths can be easily classified even when the training dataset is unbalanced using convolutional variational autoencoders for augmenting the classes with fewer samples.

For future work, we will keep working on the idea of using CNN to deal with variable length audios. The ICBHI dataset samples have similar audio lengths and it would be interesting to be able to train and predict our data without crop or resize the spectrograms no matter how long the audio is.

In addition, it would be nice to be able to detect which parts of the spectrogram emphasizes the disease in order to determine maybe new understandable symptoms by the specialists or even found multiples diseases in the same sample but at different time.

**Author Contributions:** Conceptualization, M.T.G.-O. and J.A.B.-A.; Formal analysis, M.T.G.-O. and I.G.-R.; Investigation, M.T.G.-O., I.G.-R. and C.B.; Methodology, M.T.G.-O. and H.A.-M.; Software, M.T.G.-O. and J.A.B.-A.; Validation, M.T.G.-O., I.G.-R. and C.B.; Writing—original draft, M.T.G.-O.; Writing—review & editing, J.A.B.-A., I.G.-R., C.B. and H.A.-M. All authors have read and agreed to the published version of the manuscript.

**Funding:** This study will be funded by "University of León".

**Acknowledgments:** We gratefully acknowledge the support provided by the Consejería de Educación, Junta de Castilla y León throught project LE078G18. UXXI2018/000149. U-220.

**Conflicts of Interest:** The authors declare no conflict of interest